\documentclass[sigconf]{acmart}
\AtBeginDocument{%
  }

\renewcommand\footnotetextcopyrightpermission[1]{} %
\setcopyright{none}

\usepackage{xspace}
\usepackage{flushend}

\usepackage{subfigure}

\AtBeginDocument{%
  }

\newcommand{\nc}{\textsc{N2c2}\xspace}

\newcommand{\mlm}{\textsc{mlm-bert}\xspace}
\newcommand{\mlma}{\textsc{mlmA-bert}\xspace}
\newcommand{\mlmdp}{\textsc{DP-bert}\xspace}
\newcommand{\ppmlmi}{\textsc{IPPmlm-bert}\xspace}
\newcommand{\ppmlmd}{\textsc{DPPmlm-bert}\xspace}
\newcommand{\ppmlm}{\textsc{PPmlm-bert}\xspace}

\begin{document}

\title{Towards the Anonymization of the Language Modeling}





\author{Antoine Boutet}
\affiliation{%
 \institution{INSA Lyon, Inria, CITI, UR3720}
 \city{69621 Villeurbanne}
 \country{France}}
 \email{antoine.boutet@insa-lyon.fr}
 
\author{Lucas Magnana}
\affiliation{%
  \institution{Inria, INSA Lyon, CITI, UR3720}
  \city{69621 Villeurbanne}
  \country{France}
}
\email{lucas.magnana@inria.fr}

\author{Juliette Sénéchal}
\affiliation{%
  \institution{Université de Lille}
  \city{Lille}
  \country{France}}
\email{juliette.senechal@univ-lille.fr}



\begin{abstract}
Rapid advances in Natural Language Processing (NLP) have revolutionized many fields, including healthcare. However, these advances raise significant privacy concerns, especially when pre-trained models, fine-tuned and specialized on sensitive data, can memorize and then expose and regurgitate personal information.
This paper presents a privacy-preserving language modeling approach to address the problem of anonymization of language models, and thus promote their sharing.
Specifically, we propose a Masked Language Modeling (MLM) methodology to specialize a BERT-like language model
that prevents the model from memorizing direct and indirect identifying information present in the training data.
We comprehensively evaluated our approach on several models using a medical dataset and a corpus of legal texts, and compared it to different baselines. 
Our results indicate that by avoiding memorizing both direct and indirect identifiers during model specialization, our masked language modeling schemes offer a good tradeoff for maintaining high privacy while retaining high utility.
\end{abstract}




\maketitle
\pagestyle{plain}

\section{Introduction} 

With the rise of Machine Learning (ML) and the advent of Natural Language Processing (NLP), the need for information sharing in hospitals now concerns learning models.
As pre-trained Large Language Models (LLMs) specialized and fine-tuned on specific clinical reports improve their performance for downstream NLP tasks~\cite{huang2020clinicalbertmodelingclinicalnotes,labrak2023drbertrobustpretrainedmodel,dura2022learningstructuresfrenchclinical}, many  hospitals wish to share their models.
In this context, the model fine-tuned with the specific corpus 
must not regurgitate personal and sensitive data used during its specialization (or return any other information allowing to deduce them) when it will be used once shared. 
The attack surface on models trained on personal and highly sensitive data is still poorly understood~\cite{10.1145/3531146.3533088,lehman-etal-2021-bert,duprieu:hal-04782667}.
A number of threats are related to the memorization and possible leakage of personal information used during the initial training or the fine-tuning of the model, such as 
the membership inferences~\cite{tao2025tokenlevelinformiastrongermembership} (i.e., identifying elements used during the training or the fine-tuning).

Thus, in order to share a model without risks of regurgitation of sensitive information, it is necessary to ensure that the model has not memorized personal information. 
This was recalled by the European Data Protection Board (EDPB), which issued an opinion on the development and deployment of AI models in light of the GDPR\footnote{EDPB opinion on AI models - \url{https://www.edpb.europa.eu/news/news/2024/edpb-opinion-ai-models-gdpr-principles-support-responsible-ai_en}} 
\footnote{AI Privacy Risks \& Mitigations LLMs report - \url{https://www.edpb.europa.eu/our-work-tools/our-documents/support-pool-experts-projects/ai-privacy-risks-mitigations-large_en}}. 
More precisely, the opinion therefore considers that an AI model trained on personal data cannot be considered anonymous in all cases, and that in most cases, an analysis will have to be conducted by the data controller to demonstrate that such a model is anonymous (i.e., the likelihood of regurgitation of personal information must be insignificant taking into account all the means reasonably likely to be used).

To reduce privacy risks, some practitioners exploit Named Entity Recognition (NER) to pseudonymize the  training data~\cite{speicher2024understandingmemorisationllmsdynamics}.
A NER is a model trained to recognize a set of Personally Identifiable Information (PII) that directly identify patients. 
However, pseudonymization is not anonymization~\cite{pilan2022textanonymizationbenchmarktab,xin2025falsesenseprivacyevaluating}. 
Indeed, a link between an individual and their data can be found from indirect identifying information. Therefore, these indirect identifiers must also be protected (e.g., removed) in order to ensure anonymization.
This operation cannot rely on a NER task as there is no prior information to identify these indirect identifiers, it depends on the context and the considered dataset.
Other mitigation techniques have been proposed such as learning models with Differential Privacy (DP) guarantees~\cite{dp,mckenna2025scalinglawsdifferentiallyprivate} or pruning strategies~\cite{10020711}. While these mitigation techniques increase the probability of the adversary making a wrong decision or inference, these countermeasures also significantly degrade the model's accuracy, making them difficult to use in practice. 
%




In this paper, we propose privacy-by-design language modeling approaches to address the problem of anonymizing language models. Specifically, we propose a Masked Language Modeling (MLM) methodology to specialize pre-trained masked language models (e.g., BERT-like models) 
that prevents the model from memorizing direct and indirect identifying information.
To achieve this goal, our fine-tuning approach named \ppmlm first identifies directly and indirectly identifying information in the data corpus used for the specialization. 
While direct identifiers are identified by exploiting an off-the-shelf NER model, indirect identification words are defined as words (or n-grams) used only by a single patient (which goes beyond the directly identifying words). This ensures that the memorized words were used in the data of at least two individuals (i.e., thus making the k=2 individuals indistinguishable; note that the value of k can be parameterized in our solution).
Then, the language modeling to specialize and fine-tune the model on the training data (i.e., the random masking for masked models) 
avoids using directly and indirectly identifying words. 
Therefore, the risk of regurgitation or inference of personal information during the use of the model is drastically reduced.

We comprehensively evaluated our approach on several models (e.g., BERT, RoBERTa) using both medical reports and legal texts, and considering a wide range of comparison baselines (e.g., the use of pseudonymized data for training and DP).
We quantified the impact of our solution on the resulting model's performance using multiple utility metrics (e.g., the ability to predict words and training downstream classification tasks), as well as on privacy metrics based on the model's ability to predict identifiers or through Membership Inference Attacks (MIAs).
We show that ignoring direct and indirect identifiers in the language modeling improves privacy while maintaining good utility.
To enable reproducibility of results, 
code and public data as well as other associated research artifacts are available\footnote{Code for reproducibility: \url{https://gitlab.inria.fr/aboutet1/noid-llms} }.

To summarize our contribution, this work addresses the too-often-seen limitation that relying on NER models to detect identifying information and simply removing it from the training data is enough to adequately protect privacy~\cite{xin2025falsesenseprivacyevaluating}. We believe that recognizing this limitation and proposing a solution could be useful in practice to prevent models supposedly without personal information (i.e., who have only exploited a NER) from being freely exchanged on Hugging Face\footnote{Hugging Face - \url{https://huggingface.co/}}.
Finally, our methodology preventing the memorization by the model of both direct and indirect identifiers aims to contribute to alignment with the EDPB's opinion regarding AI models trained with personal data that must not regurgitate personal information to ensure their anonymous nature.

\section{Background and related work} 
\label{sec:sota}


\subsection{Natural Language Processing}
\label{sec:NLP}

Natural language processing (NLP) is the process of understanding and processing textual data using ML models. 
From the advent of the Transformer~\cite{transformer}, 
the architecture of the neural network 
consists of an encoder-decoder with a parallel computational scheme that uses positional encoding and various attention mechanisms~\cite{illustrated}.
Two trends then followed with BERT models~\cite{devlin2019bert} that focus on the encoder part of the Transformer, and GPT models~\cite{openai2024gpt4technicalreport} that use the decoder. The former learns very complex word embeddings and are very effective for classification tasks, while the latter are generative models trained to imitate human expression and useful to create chatbots for instance. 
These models are named foundation models, they are pre-trained on huge unlabeled datasets. It is then possible to specialize and fine-tune such models on specific data through language modeling, and even learn new tasks by adding just a few layers to the model (in the case of BERT models). 
%
%
For example, RoBERTa~\cite{liu2019roberta} is a much larger version of BERT, or distilBERT~\cite{sanh2020distilbert} uses knowledge distillation to produce a smaller model. Models in other languages have also emerged, such as CamemBERT~\cite{camembert} for French, or specialized on specific data corpus such as ClinicalBERT~\cite{peng2019transferlearningbiomedicalnatural} or DrBERT~\cite{labrak2023drbertrobustpretrainedmodel} which have been fine-tuned with clinical reports. 




Language modeling consists of predicting the words in a document. 
On the one hand, Causal Language Modeling (CLM) consists of predicting words sequentially, from left to right. 
This is the most common approach used in text generation with a GPT-like model.
On the other hand, Masked Language Modeling (MLM) is when the prediction task is done anywhere in the text and in any order, in other words, it is chosen randomly. 
This is the approach mainly used for language understanding tasks with a BERT-like model.
In both cases, the prediction is based on the context. This means that the words before the mask (in the CLM case) or the words around the mask (in the MLM case) allow the model to identify what the word to fill could be.
Performing this language modeling task on texts from a specific corpus allows it to be specialized on that one, and improves the model's performance for downstream NLP tasks~\cite{huang2020clinicalbertmodelingclinicalnotes,labrak2023drbertrobustpretrainedmodel,dura2022learningstructuresfrenchclinical}.
Our work focuses on MLM for the construction of BERT-like models that will be used for downstream NLP tasks, a widespread use case in hospitals.





In order to exploit or share their patients' information for research purposes while ensuring patient confidentiality, hospitals have begun to exploit language models to de-identify clinical reports~\cite{8904544,tannier2023development,richardhal-04139391,hartman2020customization,tchouka2022deidentification}.
De-identification of text-based clinical reports involves removing or replacing Personally Identifiable Information (PII) from electronic health records.
The set of PII to consider is recommended by Data Protection Authorities (DPA)~\cite{dpa-cnil-pii} and includes direct identifiers such as names, hospitals, birth dates, and patient numbers. 
However, no studies have focused on indirect identifiers.

\subsection{Privacy Leakages}
\label{sec:privacy4nlp}

By being exposed to a large amount of potentially sensitive data during its training or specialization, a language model can memorize it and then regurgitate or reveal it once deployed or shared~\cite{carlini2019secretsharerevaluatingtesting,nasr2023scalableextractiontrainingdata,counterfactual}.
%
%
However, defining memorization for language models is challenging, and many existing definitions and notions have been proposed depending on whether the memorization concerns copyrighted content or personal and sensitive content.
In relation to privacy, we can notably cite extractable memorization and membership inference.\\
\textbf{Extractable memorization} is a type of attack that aims to use the model to infer information from the original data~\cite{privacy-survey}. This attack mainly concerns text generation models, such as GPT. These models are trained to produce text based on what they have seen during training. However, the model is not expected to be a basic parrot and repeat exactly the sentences it has seen.  This is of particular concern if the repeated data are sensitive~\cite{gpt2}.  
In~\cite{extract}, the term $k-extractability$ is used to refer to the sequences that can be extracted from the model when an input sequence of length $k$ is requested. The lower the $k$, the easier it is to extract the sequence. We therefore expect a model to have the highest possible $k$ on private queries.
This measure, however, does not capture regurgitations that are not perfect, which can lead to the illusion of no extractable memory.
Compressible memorization~\cite{schwarzschild2024rethinkingllmmemorizationlens} extends this definition by evaluating how short the minimal requested sentence (or prompt) that elicits the sequence.\\




\textbf{Membership inference attacks}~\cite{carlini2022membership} (MIAs) 
aim to infer whether a specific data was used in the training data of a target model. 
%
There are different techniques that can be used to perform a MIA attack depending on if the adversary has access or not to the model parameters (i.e., white-box versus black-box access), or access to a ground-truth
subset of member and non-member samples. One technique consists to analyze the loss of member and non member samples~\cite{8429311}, another one is to use multiple shadow models~\cite{shokri2017membership,10.1145/3548606.3560675} trained to mimic the behavior of the target model on an auxiliary dataset.
An adversarial model is then trained to infer membership from the loss or from shadow models.
Another method~\cite{counterfactual} is based on comparing the performance of the target model trained on a dataset with a specific input, with a second model trained without it.
As ML models are supposed to learn general information, one piece of data (even if it is rare, an outlier, or mislabeled) is not supposed to be memorized and should not significantly change the model's performance. By repeating this operation many times with different subsets, it is possible to identify counterfactually memorized data.
%
%
%
Although membership inference attacks have been used to quantify memorization risks~\cite{jagannatha2021membership,mireshghallah2022quantifying,10020711,duan2024membership,10.1145/3658644.3690194}, they may not be applicable in a practical scenario due to the high-dimensional input space (i.e., too expensive in terms of calculation). 
Being able to evaluate privacy at low-cost is important and challenging~\cite{zarifzadeh2024lowcosthighpowermembershipinference}.
Indeed, using MIAs that require training multiple models is intractable for LLMs.





\subsection{Mitigation strategies}
\label{sec:mitigation}

To reduce privacy risks, many works~\cite{xin2025falsesenseprivacyevaluating}
only exploit Named Entity Recognition (NER) to remove the direct identifiers in the training data.
Empirically, \cite{chen2022relaxlossdefendingmembershipinference} proposed to relax the loss in order to reduce the distinguishability between the training and testing loss distributions, and~\cite{chen-etal-2024-learnable} proposed a method to localize specific neurons responsible for memorizing PII in LLMs through adversarial training and then disable these neurons.
Theoretically, Differential Privacy~\cite{dwork2006differential} (DP) is a mathematical property to protect privacy. The most popular method to apply it in machine learning is DP-SGD: Differentially-Private Stochastic Gradient Descent~\cite{dp}. The idea is to apply DP during the training phase by clipping gradient updates and adding centered noise at each step. More formally, the probability that a model guesses the correct output for a given input must not increase too much each time the model sees that data: $\forall (x,y),\, \log P(M_D(x)=y)< \epsilon \log P(M_{D+x}(x)=y),$ where $x,y$ represents data and its label, $M_D$ a model trained on dataset D and $\epsilon$ the ~\textit{privacy budget}. The lower $\epsilon$ is, the more private the model is.
DP is known to significantly decrease the accuracy of the model~\cite{jagannatha2021membership} and privacy budget management is difficult.
Libraries are emerging to apply DP to language models~\cite{dp-transformers}.
In~\cite{yu2022differentiallyprivatefinetuninglanguage}, A differentially private fine-tuning version of pre-trained generative models have been proposed, however the regurgitation of identifiers has not been considered.
To improve the utility, \cite{li2024finetuninglanguagemodelsdifferential} has proposed to adapt the noise allocation.

\begin{figure}[t]
    \centering
    \includegraphics[scale=0.05]{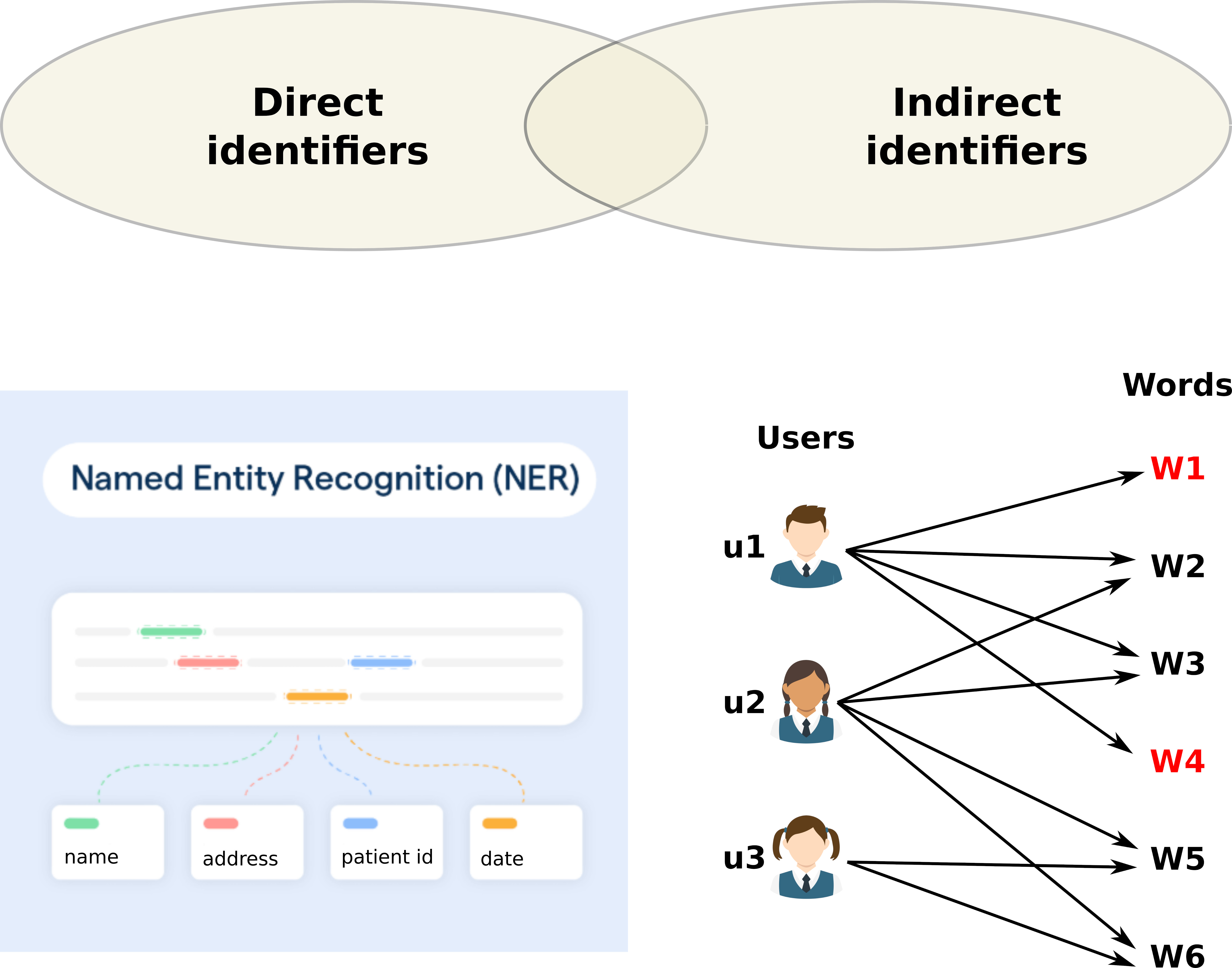}
    \vspace{1mm}
    \caption{Detection of the identifiers: a off-the-shelf NER model can detect the direct identifiers while by constructing a bipartite graph between individuals and the words used in their documents, we can easily identify indirectly identifiers (i.e., words pointed to only by one individual).}
    \label{fig:bipartite}
\end{figure}

\section{Privacy Auditing} 
\label{sec:problem}

On the basis of the opinion 28/2024 of the EDPB related to the processing of personal data in the context of AI models, we are considering a privacy audit performed by the system designer to ensure the anonymous nature of the model.
Specifically, this means that we are considering an attack model where the adversary knows the dataset used for the model specialization and wants to ensure that the likelihood of regurgitation or leakage of personal information contained in this dataset (i.e., both all direct and indirect identifiers) is insignificant.

%
%
%
%
%
%
Direct identifiers include information associated with a natural person such as first and last name, a telephone, or a social security number. 
However, direct identifiers fail to detect all terms that can lead to distinguishing an individual.
This distinguishing nature is captured by indirect identifiers.
Indirect identifiers are information that is uniquely used in an individual's data compared to the rest of the population.
For example, if the term vertebra is used only in a patient's data, this term is indirectly identifying compared to the rest of the corpus. In other words, the individual is distinguishable from others with the use of this term.


While the direct identifiers can be detected by an off-the-shelf NER model\footnote{e.g., EDS-Pseudo trained from Clinical Data Warehouse - \url{https://github.com/aphp/eds-pseudo}}, indirect identifiers have to be detected. To achieve this goal, an analysis of the data corpus must be carried out. 
Figure~\ref{fig:bipartite} depicts a bipartite graph that links the individuals represented in the data corpus to the words used across the entire dataset.
More formally, it is a Bipartite Graph $G=(V,E)$ where the vertices of this graph are the users and words of the dataset $V=(U \bigcup I)$. The use of a word in a user's data is represented in $G$ by an edge $(u, i) \in E \subseteq U \times I$.
In this example, we can see that the words $w1$ and $w4$ are only used in the data of individual $u1$.
Note that a direct identifier can also be a word used only in an individual's data (i.e., an indirect identifier), this is the typical case of a last name. Note also that an indirect identifier can be composed of multiple terms (i.e., n-grams).

These identifying data (direct and indirect) can pose a confidentiality issue if they are memorized by the model before sharing it.
Indeed, an identifier memorized by the model due to its specialization and proposed (i.e., predicted during a masking phase) or regurgitated (i.e., proposed to complete a sequence) during model exploitation would betray its use during training (i.e., it would represent a membership inference).

\section{Privacy-Preserving Masked
Language Modeling (PP{\scriptsize{MLM}})}
\label{sec:pplm}

In order to prevent the model from memorizing personal information, the privacy-preserving masked language modeling that we propose ensures that the choice of masks used during the specialization of a pre-trained model avoids directly and indirectly identifiers. 
To achieve that, this solution relies on preprocessing that is responsible for identifying both the direct and indirect identifiers. 

\subsection{Preprocessing: Building a Blacklist}
\label{sec:pp-preprocessing}


The preprocessing step focuses on identifying identifiers that the model must not to memorize.
First, we use an off-the-shelf NER model to directly identify identifying words associated with certain categories. This NER model can be adapted to include more or less categories depending on the field of specialization. Then, we build the bipartite graph $G$ between the individuals of the corpus and the words used in their documents in order to identify indirect identifiers (similar to Figure~\ref{fig:bipartite}). To do this, once the graph is built, we identify the words that are only linked to one individual. Finally, all these (directly or indirectly) identifiers are kept in a kind of blacklist that is used during the following masked language modeling.

If we make an analogy with the notion of k-anonymity~\cite{sweeney2002k}, which defines that a data is anonymous if it is used by at least k different users, all the words used as masks to specialize the model in the language modeling step are at least used by two people (i.e., k=2). This makes the task of an adversary wishing to infer the membership of an individual more difficult.
We can easily change the exploitation of the bipartite graph to consider either a larger value of k for the identification of indirectly identifying information (e.g., words that are linked to 1 or 2 users to have a k=3), or the identification of identifying n-grams (i.e., identifiers composed of multiple terms).  
Additionally, it is important to note that the cost of this preprocessing is reduced.




\subsection{Masked
Language Modeling: avoiding sensitive words}
\label{sec:ppmlm}

Compared to the standard masked language modeling scheme that randomly chooses terms to mask during supervised learning, our privacy-preserving masked language modeling (\ppmlm) ensures that directly or indirectly identifying terms are never masked by leveraging the blacklist constructed during the preprocessing step. More precisely, the choice of the mask is made randomly among the words that are not in the blacklist.
However, the identifying words (i.e., the words kept in the blacklist) can be used in the context of a mask to be predicted. Their use improves the prediction of masked terms without introducing a significant risk of memorization leading to regurgitation.
We evaluate and compare the impact on the utility and privacy tradeoff of memorizing direct identifiers as a mask or as context by using a baseline that pseudonymizes training data before specialization (i.e., direct identifiers are removed from the training data and cannot be used as context).

\section{Evaluation}
\label{sec:results}

This section presents a comprehensive evaluation of our privacy-preserving masked language modeling scheme.
We show that memorization of identifiers can lead to the membership inference of individuals whose data were used in training. Furthermore, we show that protecting the direct identifier only through pseudonymization is not enough to protect the model against a risk of memorizing indirect identifiers. 
We also show that the most leaked identifying information is that which is most repeated in the training data. 

\subsection{Experimental Setup}


\paragraph{Datasets.}
To conduct our experiments on health data, we considered the \nc (National NLP Clinical Challenges) datasets~\cite{DBLP:journals/jamia/HenryBFSU20}. 
More precisely, we leverage two datasets gathering medical discharge summaries (i.e., English free text).
In the first one~\cite{10.1197/jamia.M2444}, 928
records were annotated for replacing all authentic Personally Identifiable Information (PII) 
with realistic surrogates. These PIIs fall into different categories: six of the seventeen textual PII categories listed by HIPAA (only these six categories appear in the data: patients, locations, dates, IDs, phone numbers, and ages), and two additional categories, doctors and hospitals, resulting in eight PII categories in the dataset.
In addition, a subset of 502 anonymized discharge summaries from this first dataset was annotated with the patient's smoking status~\cite{10.1197/jamia.M2408}.
In the second dataset~\cite{10.1197/jamia.M2408}, almost 1250 records were annotated by two obesity experts. The experts were asked to classify fifteen frequently occurring obesity co-morbidities as Present, Absent, Questionable, or Unmentioned based on explicitly documented information in the patient records. 
We leverage these datasets to assess the utility degradation of our privacy-preserving language modeling through a downstream classification task (i.e., classification of obesity comorbidities, and smoking status).

\paragraph{Models Architecture.}

We consider three masked language models: the BERT base cased~\cite{devlin2019bert} (composed of 12 attention heads, 768 hidden dimensions, 110M parameters), the RoBERTa Large~\cite{DBLP:journals/corr/abs-1907-11692} (composed of 24 attention heads, 1024 hidden dimensions, 340M parameters), and the DistilRoBERTa~\cite{DBLP:journals/corr/abs-1910-01108} (composed of 6 attention heads, 768 hidden dimensions, 66M parameters) 
from Hugging Face\footnote{
Hugging Face - \url{https://huggingface.co/}}. 
These different models allow us to assess the impact of model size.





\paragraph{Metrics.}

We consider utility and privacy measures in our evaluation to better assess the trade-off between protecting training data and the performance of the model in language modeling. 
%
For the utility evaluation, we 
quantify the ability of the models to predict masked terms well (measured by the prediction accuracy), and the model’s performance in classification by creating a downstream task (measured by the F1-Score metric on the prediction of the obesity co-morbidities, or smoker status).
%
%
To assess the model' privacy, we study its ability to predict identifiers (direct and indirect) contained in the dataset used to specialize the model. An individual's presence is revealed as soon as at least one of their identifiers is predicted or regurgitated by the model, whether this prediction is in the correct place in the original text or elsewhere.
Moreover, these identifiers can be predicted once or multiple times.
We use the metric \textit{Privacy} to measure the ability of a model to predict any identifying term (i.e., direct or indirect identifier) at least once. A Privacy=1 means that no identifiers are predicted 
by the model, and a Privacy=0 means that all identifiers are leaked. 
Although this metric does not account for multiple predictions, we also evaluate the number of times an identifier is predicted based on its frequency in the training data.
%
We also considered MIAs to measure an adversary's ability to correctly predict whether a patient's reports were used in model fine-tuning.
We designed a cost-efficient Membership Inference Attack (MIA) that relies on the count of predicted indirect identifiers. Each patient receives a score reflecting how many of their indirect identifiers were inferred. The goal of the attack is to identify an optimal threshold that balances maximizing true positives while minimizing false positives. In this scenario, the adversary is able to construct the same bipartite graph as the one used in the fine-tuning (i.e., the adversary has the data used to fine-tune the model, which corresponds to a privacy audit carried out by the designer on his own model).



\paragraph{Comparative Baselines.}
We compare our privacy-preserving masked language modeling scheme  (named \ppmlm) to different comparative baselines. 
First, we consider a classical specialization of BERT models (named \mlm).
Next, we consider a baseline that performs the same specialization but using a pseudonymized dataset for the fine-tuning (baseline named \mlma).
We also consider two variants of our solution where only direct or indirect identifiers are protected from memorization (baselines named \ppmlmd and \ppmlmi, respectively).
Finally, we consider a state-of-the-art baseline using a differentially private fine-tuning of pre-trained language models integrating the Opacus library\footnote{Opacus - \url{https://opacus.ai/}} (baseline named \mlmdp).
%
%
%
We also considered~\cite{chen-etal-2024-learnable} which proposes to localize specific neurons responsible for memorizing PII in LLMs and then disable them. However, no repository is available and we were unable to reproduce their results.

\paragraph{Methodology.}


This section provides the methodological details we followed to perform our evaluations.
First, when datasets are pseudonymized before being used for model fine-tuning (i.e., in the case of \mlma), direct identifiers are replaced by the term "X" in the text.
Then, BERT models were trained by randomly replacing a subset of tokens from the sequence by a masking token, and asking the model to predict them using cross-entropy loss. In our setting, 15\% of the words are randomly selected.
In the case of our solution, identifying words are excluded from this random selection.
If a word chosen to be masked consists of several tokens, all of them are masked. At each epoch, 15\% of non-identifying words are masked.

\begin{figure}[t]
    \centering
    \includegraphics[width=0.39\textwidth]{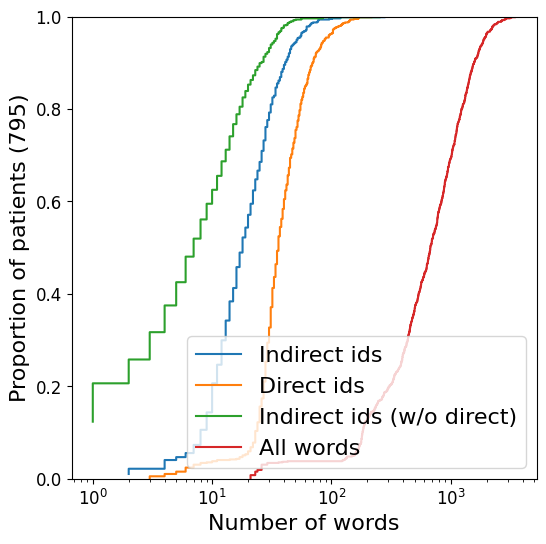} 
     \caption{Cumulative distribution of the number of identifiers (both direct and indirect ones), and words per patient: half of the patients have more than 20 indirect identifiers, and almost all patients have at least 3 indirect identifiers.}
    \label{fig:cdf_ind_ids}
\end{figure}

All measurements were performed after 4, 8, 16, and 32 epochs to observe their evolution. 
Given the size of our datasets, 16 and 32 epochs correspond to overfitting. We deliberately pushed the training to an overfitting state in order to evaluate this scenario as well.
We use batches of size 8, with each sequence in the batch containing 512 tokens.
For language modeling, the learning rate starts at 1e-4 and decreases linearly until the end of finetuning. For classification, we took the finetuned model after 32 epochs and finetuned it again for 6 epochs while freezing all the BERT layers (to not bias the model and introduce other leakage) except the last one and the classification layer following the methodology described in this study~\cite{sun2020finetuneberttextclassification}. The learning rate this time starts at 0, 10\% of the total number of warmup steps with the learning rate increasing until reaching 2e-5 then decreasing linearly on the remaining 90\% steps.
For DP experiments, we use the same methodology as described in~\cite{li2024finetuninglanguagemodelsdifferential} using $\epsilon=8, C=10, \delta=1e-5$.
All the computation has been parallelized on a hybrid GPU/CPU computing farm.\\

\subsection{Identifiers Analysis}
\label{sec:idanalysis}

In this section, 
we first analyze the distribution of direct and indirect identifiers in the \nc dataset, as well as the number of words in each patient’s reports (Figure~\ref{fig:cdf_ind_ids}).
We can see that almost all patients have indirect identifiers and half of them have more than twenty indirect identifiers. 
We can also see that direct identifiers are about twice as numerous as indirect identifiers.
Finally, a fraction of indirect identifiers (i.e., words exploited in the documents of a single patient) are also direct identifiers.

\begin{figure}[t]
    \centering
    \includegraphics[width=0.39\textwidth]{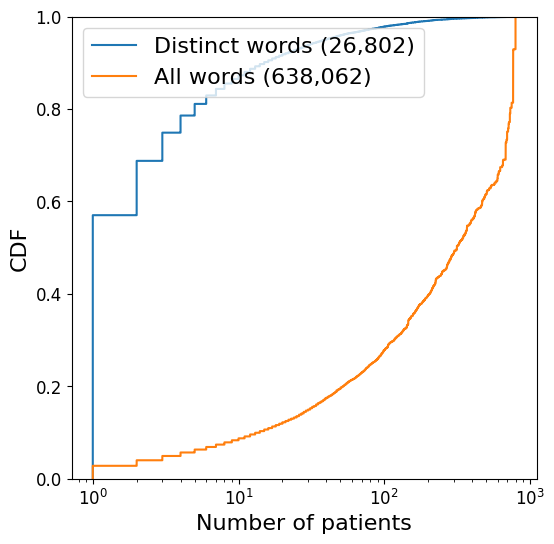} 
     \caption{Cumulative distribution of the number of patients who share words in their reports: 60\% of distinct words in the corpus are used by only one patient and therefore represent indirect identifiers, however these words represent only 3\% of the total number of word occurrences in all reports.}
    \label{fig:prop_ids_}
\end{figure}

Figure~\ref{fig:prop_ids_}, in turn, shows 
the cumulative distribution of the number of patients who share words in their reports. The plot distinguishes distinct words from the total number of words by including repetitions.
The distribution shows 
that 60\% of distinct words in the corpus are used by only one patient and therefore represent indirect identifiers.
However, the use of these words represents less than 3\% of the total occurrences of words present in all reports. In other words, the 40\% of words used in reports of multiple patients represents more than 97\% of the total volume of words used.

%


 

\begin{figure}[t]
    \centering
    \includegraphics[width=0.49\textwidth]{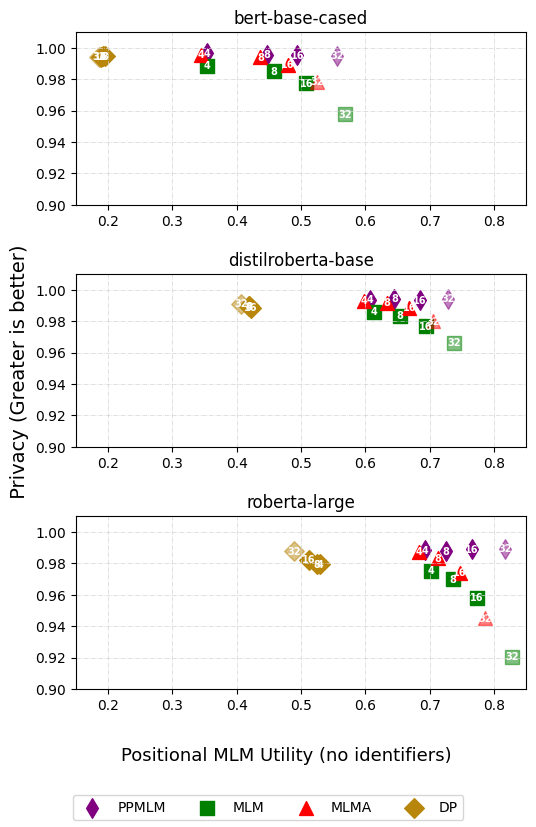}
 \label{fig:mlmprivacyN2C2}
 
         \caption{Utility and privacy evaluation: by avoiding memorizing both direct and indirect identifiers during model specialization, our \ppmlm solution offers the best tradeoff in maintaining high privacy while retaining high utility.}
    \label{fig:mlmprivacy}
\end{figure}

 

 

 

\subsection{Utility and Privacy Tradeoff} 
\label{sec:eval-mlm}


We now evaluate the utility and privacy tradeoff for our masked language modeling scheme.
Figure~\ref{fig:mlmprivacy} illustrates this tradeoff for \ppmlm and other comparative baselines on the three considered BERT-like models. 
First, the results show that for all models that were specialized without a specific mechanism to prevent the model from memorizing identifiers (i.e., classical \mlm), privacy is significantly degraded, from 0.97 after 4 epochs to 0.92 after 32 epochs for RoBERTa Large.
This decrease in privacy means that the model learns more and more identifiers over the epochs and is led to predict them more along the training.
Second, training using a pseudonymized dataset (i.e., leveraging a training dataset without direct identifiers, \mlma) slightly improves the level of privacy (from 0.92 to more than 0.94 after 16 epochs for RoBERTa Large) but reduces the utility (from 0.83 to 0.79 after 16 epochs for RoBERTa Large). This reduction in utility comes from a context used for prediction that does not benefit from direct identifiers.
Third, \mlmdp offers a good level of privacy (around 0.98 for all models) 
but drastically impacts the utility of the prediction (nearly 0.20, 0.40, and 0.50 of utility for BERT Base, DistilRoBERTa, and RoBERTa Large, respectively).
Finally, our \ppmlm solution offers the best utility and privacy tradeoff for all models, where the privacy remains stable around 0.99 regardless of the number of epochs, meaning that the model learns almost no identifiers, either directly or indirectly, during fine-tuning.
The difference in privacy between \ppmlm and \mlma is due to the prediction of indirect identifiers for \mlma, as direct identifiers have been removed from the training dataset.
\ppmlm also offers a 
high utility due to the wealth of information (i.e., using all terms, including identifiers) used in the context to predict a term.
The results also show that model size mainly influences its utility (the larger, the better). However, the RoBERTa Large model tends to predict identifiers more frequently than the two smaller models.

In Appendix~\ref{sec:annexe}, we examine the behavior of \ppmlm when only direct or indirect identifiers are protected from memorization (i.e., using the \ppmlmd and \ppmlmi baselines). 
The impact of parameter $k$ is analyzed in Appendix~\ref{sec:k} while Appendix~\ref{sec:impactngrams} analyzes the impact if indirect identifiers are composed of multiple terms (i.e., n-grams). 
Appendix~\ref{appendix:legal}, in turn, assesses the generality of our solution by using a corpus of legal texts from judgments.

\begin{figure}[t]
    \centering
-    \includegraphics[width=0.49\textwidth]{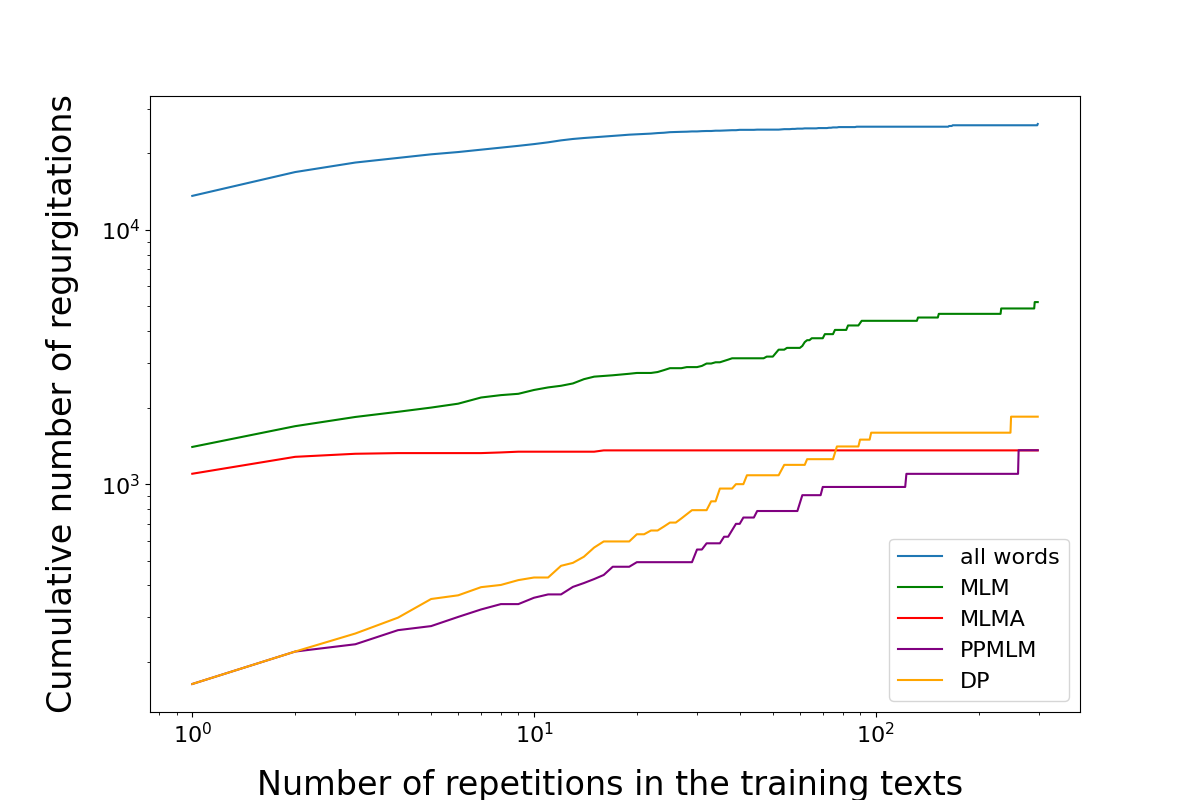}

     \caption{Regurgitation (RoBERTa LARGE): the most frequently regurgitated identifiers are the most repeated words in the training dataset.}
    \label{fig:mlmprivacyN2C2regurgitation}
\end{figure}

Regarding the characterization of regurgitation, Figure~\ref{fig:mlmprivacyN2C2regurgitation} presents the cumulative distribution of identifier regurgitation as a function of their repetition in the training set.
First, we observe that identifiers can be repeated up to 200 times in the training set.
Next, the results illustrate the effect of our mitigation strategy on the number of regurgitation. For \ppmlm, the number of regurgitation is drastically reduced compared to an unprotected solution (\mlm) and a solution trained with pseudonymized data (\mlma). This privacy leakage reduction corresponds to the area between the curves.
Lastly, the results show that the most frequently predicted (i.e., regurgitated) identifiers correspond to the most repeated ones in the training set.
Note the low number of repetitions in the \mlma solution, which shows that repetitions are more common with direct identifiers than indirect ones.
And finally, the results show that \mlmdp only slightly increases the number of regurgitations compared to \ppmlm, mainly on repeated terms in the training dataset.

\begin{table}[t]
\centering
\begin{tabular}{ |c|c|c| } 
\hline
& \nc Obesity & \nc Smokers  \\
\hline
\ppmlm & 0.86 & 0.60 \\  
\mlm & 0.86 & 0.60 \\ 
\mlma & 0.86 & 0.60 \\ 
\hline
\end{tabular}
\vspace{2mm}
\caption{Utility (RoBERTa LARGE) assessment through a downstream classification task (after 6 epochs): all baselines provide the same level of F1-Score.}
\vspace{4mm}
\label{fig:classificationMLM}
\end{table}

We also evaluate the utility of the models through two classification tasks applied to another dataset \nc.
This classification task aims to predict the patient's obesity co-morbidities and the patient's smoker status, and the classification models are trained as a downstream task from the model specialized on the initial dataset (i.e., annotated for replacing all PII).
%
Table~\ref{fig:classificationMLM} reports the F1-Score of the classification model after 6 epochs.
Results show that all models (i.e., \ppmlm, \mlm, and \mlma) provide the same level of F1-Score for both classification tasks.
This means that by preventing the memorization of identifiers, our approach has no impact on the accuracy of the classification.

\begin{figure*}
  \begin{minipage}[b]{.32\linewidth}
    \centering
    \includegraphics[width=1\textwidth]{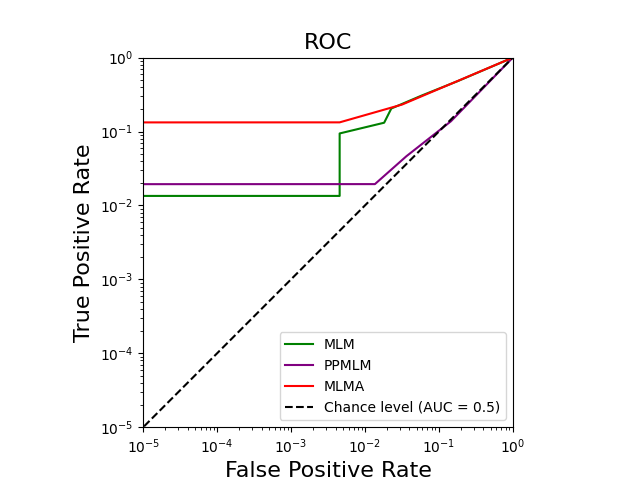}
    \texttt{BERT BASE-CASED}
  \end{minipage}
  \hfill
  \begin{minipage}[b]{.32\linewidth}
    \centering
    \includegraphics[width=1\textwidth]{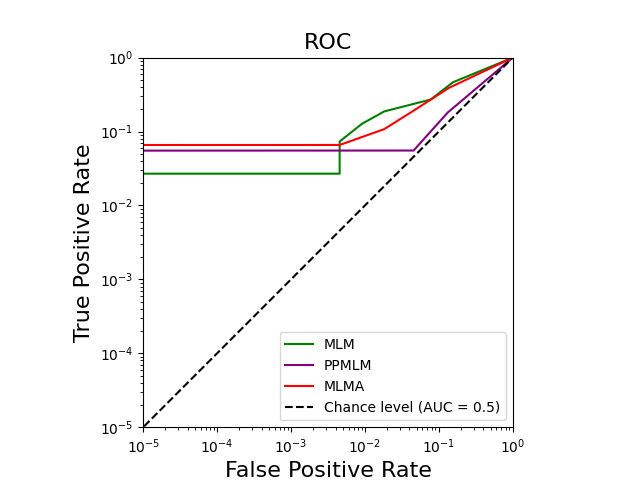}
    \texttt{DISTIL-RoBERTa}
  \end{minipage}
  \hfill
  \begin{minipage}[b]{.32\linewidth}
    \centering
    \includegraphics[width=1\textwidth]{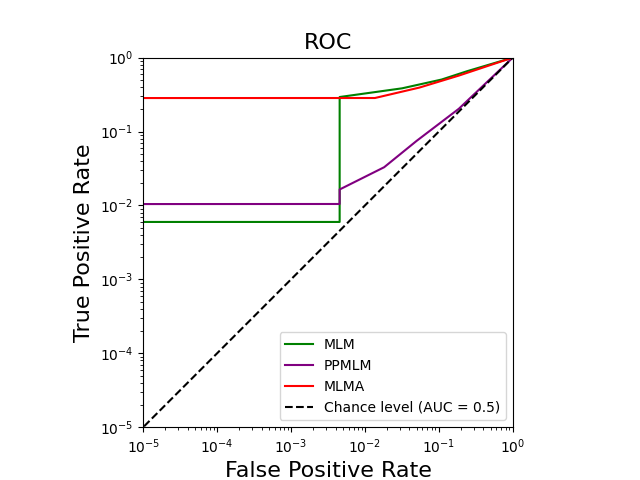}
    \texttt{RoBERTa LARGE}
  \end{minipage}
  \caption{ROC curve for the Membership Inference Attack with log-log scaling to emphasize the low-FPR regime: \ppmlm drastically reduces the accuracy of the membership prediction.}
  \label{fig:mia}
\end{figure*}

\subsection{Impact of Membership Inference Attacks}
\label{sec:impactmia}

To evaluate the accuracy of the membership inference attacks, we conducted a ROC (Receiver Operating Characteristic) analysis which consists of comparing the positive predictions (i.e., membership guesses) over negative (i.e., non-membership) predictions. Specifically, we consider the true-positive rate (TPR) at low false-positive rates (FPR) for an attack to infer membership from the prediction of indirect identifiers conducted on a fine-tuned BERT model. The ROC curve (Figure~\ref{fig:mia}) depicts the AUC (Area Under the Curve) for BERT, DistilBERT, and RoBERTa models specialized without protection (i.e., \mlm), specialized models from pseudonimized data (i.e., \mlma), and specialized models following our scheme (i.e., \ppmlm). The dashed line shows AUC=0.5 which corresponds to random guess.
Table~\ref{fig:fpr}, in turn, reports for all models the true-positive rate at three different false-positive rates (i.e., 0.1\%, 1\%, and 10\%).
Results for RoBERTa Large indicate for instance that at a low false-positive rate (FPR of 1\%), the true-positive rate reaches approximately 31\% for \mlm, 26\% for \mlma, and drops around 2\% for \ppmlm. The AUC for \ppmlm is substantially lower than for both \mlm and \mlma, demonstrating its stronger resistance to our MIA. Overall, our privacy-preserving language modeling framework effectively protects over 98\% of patients from membership inference based on indirect identifiers. These experiments also demonstrate that even with a low-cost MIA without shadow models, it is possible to infer the membership of 26\% of patients from a pseudonymized model with low error rates, increasing to 31\% for models without privacy protection.
Finally, the results show that the risk of membership inference varies with the size of the model. For example, for a low false positive rate (FPR of 1\%), the risk of membership inference in \mlm decreases from 31\% for Roberta Large (340 million parameters) to 10\% for BERT (110 million parameters).

\begin{table}[t]
\centering
\begin{tabular}{ |c|c|c|c| } 
\hline
\textbf{FPR@} & \textbf{0.1\%} & \textbf{1\%} & \textbf{10\%} \\
\hline
\hline
& \multicolumn{3}{c|}{BERT} \\
\hline
\ppmlm & 1.25\% & 1.74\% & 9.95\% \\  
\mlm & 1.22\% & 10.91\% & 36.62\% \\  
\mlma & 8.87\% & 15.28\% & 37\% \\  
\hline
\hline
& \multicolumn{3}{c|}{DISTIL RoBERTa} \\
\hline
\ppmlm & 2.46\% & 3.08\% & 13.46\% \\  
\mlm & 2.69\% & 13.3\% & 32.81\% \\  
\mlma & 4.59\% & 8.25\% & 31.35\% \\  
\hline
\hline
& \multicolumn{3}{c|}{RoBERTa LARGE} \\
\hline
\ppmlm & 0.58\% & 2.3\% & 12.24\% \\  
\mlm & 0.6\% & 31.09\% & 49.34\% \\  
\mlma & 21.5\% & 26.63\% & 45.54\% \\  
\hline
\end{tabular}
\vspace{2mm}
\caption{Inference of membership at different false-positive rates: \ppmlm effectively protects patients from membership inference based on indirect identifiers.}
\label{fig:fpr}
\end{table}

\section{Legal Discussion}

The language modeling scheme presented here avoids the model from memorizing direct and indirect identifying information present in the training data (including medical datasets and sensitive data) used for the specialization of off-the-shelf foundation LLMs.
%
%
%
%
On the basis of article 9 of the GDPR and on the opinion 28/2024 of the EDPB on certain data protection aspects related to the processing of personal data in the context of AI models\footnote{\url{https://www.edpb.europa.eu/system/files/2024-12/edpb_opinion_202428_ai-models_en.pdf}} 
as well as the AI Privacy Risks \& Mitigations Large Language Models (LLMs) report\footnote{\url{https://www.edpb.europa.eu/our-work-tools/our-documents/support-pool-experts-projects/ai-privacy-risks-mitigations-large_en}}
, it is possible to say the fine-tuning of the LLM does not give birth to an fully anonymised model, due to the fact that the full anonymisation of the model can only be reached by an anonymisation process during the data preparation and preprocessing and the training phases of the foundation model\footnote{AI Privacy Risks \& Mitigations LLMs report, p. 25 et p. 33}.
Due to the fact that the model, before fine-tuning, was not, from the outset, anonymised, what is obtained is anonymization of the secondary learning data set used during fine-tuning, but not anonymization of the set of personal data on the basis of which the foundation model learned beforehand.
In other words, a lot of personal data can be regurgitated by the fine-tuned model, even if it is not the health data on the basis of which the fine-tuning took place.
Furthermore, interactions between the training data of the foundation model and the data used for fine-tuning are not controlled and may give rise to hallucinations.
Consequently, the requirement set out in the EDPB opinion of an "insignificant probability of personal data regurgitation" to qualify the fine-tuned model as anonymous is not met (Opinion 28/2024 on certain data protection aspects related to the processing of personal data in the context of AI models).
Consequently, this means that the fine-tuned model will have to respect the conditions set out by the EDPB, in the three scenarios in the presence of a re-use described in their opinion\footnote{EDPB Opinion 28/2024 on certain data protection aspects related to the processing of personal data in the context of AI models, p. 4.}.

\section{Limitations and Conclusion}
\label{sec:limitations}

%
%

The language modeling scheme presented here prevents the model from memorizing direct and indirect identifiers present in the training data (including medical datasets and sensitive data) used for the specialization of pre-trained foundation LLMs. 
This work represents a practical step towards protecting the training data against privacy leakage through the fine-tuning of pre-trained LLMs.
In addition, by preventing the memorization of direct or indirect identifiers linked to individuals, our methodology aims to contribute to alignment with the opinion 28/2024 of the EDPB related to AI models trained with personal data and article 9 of the GDPR, and consequently facilitates model sharing. 

There are, however, some limitations in several areas.
Our solution is rather a practical step, without formal guarantees, towards a better behavioral alignment with privacy and governance norms, 
than novelty which is rarely operationalized in connection to real-world governance frameworks (e.g., GDPR).
Although the study and results have been validated on one dataset, this dataset is small in size.
A validation on a larger dataset requires more computational resources and is left as future work.
We also do not consider the case of an attacker having only partial knowledge of the individuals present in the training data; this threat model is orthogonal to our work which only considers privacy auditing to enforce responsible LLM behavior before sharing its model (i.e., the auditor knows the training data and wants to ensure that it will not be regurgitated).


Another limitation is that the identification of direct identifiers of our privacy-preserving scheme relies on a NER, and the quality of this NER can impact the performance of the protection of our solution. This limitation is common to all pseudonimization schemes relying on a NER and affects both the protection itself by missing identifying words, and the privacy metrics by not counting certain leaks of personal information. However,
our solution is not impacted by misspelled and/or foreign words.
Indeed, a common problem of NER in free text is the management of misspelled and/or foreign words that cannot be found in dictionaries.
These words can be identifiers and not be protected.
It is worth noting that the advantage of our solution is that these words will be identified as indirectly identifiers in our solution and will not be memorized.


It is also worth noting that the cost of our preprocessing step (i.e., building the bipartite graph) is significantly lower than using NER and fine-tuning the model.
For example, when fine-tuning a BERT model (with 64 epochs) using a large dataset (50,000 documents), building the bipartite graph takes 0.3\% of the overall fine-tuning time, while using NER takes 7\%.


\section*{Acknowledgement}
This work has been supported by the ANR 22-PECY-0002 IPOP (Interdisciplinary Project on Privacy) project of the Cybersecurity PEPR.

\bibliographystyle{ACM-Reference-Format}
\bibliography{sample}

\newpage
\appendix




\newpage

\section{Protection only of direct or indirect identifiers}
\label{sec:annexe}



\begin{figure}[t]
\includegraphics[width=0.45\textwidth]
{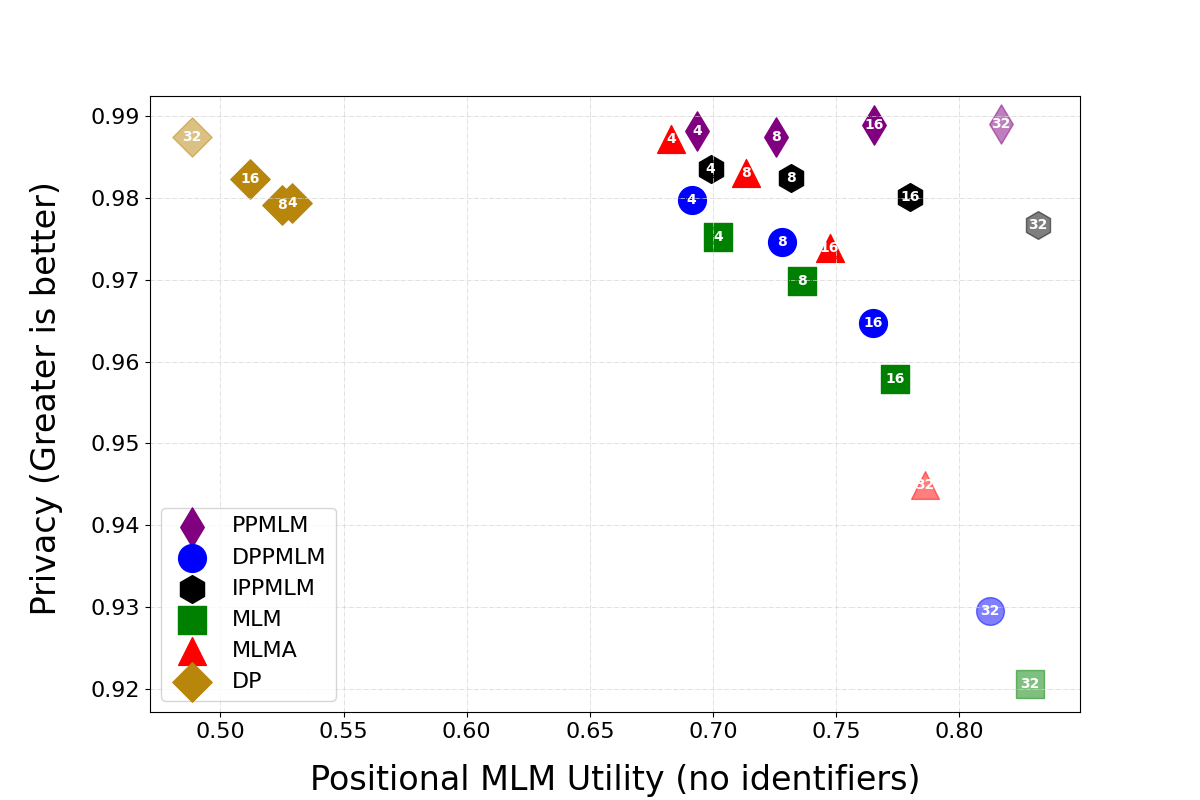}
\caption{Behavior of \ppmlm when only the direct or indirect identifiers are protected (RoBERTa LARGE): the larger number of distinct indirect identifiers than direct identifiers implies that the risk of leakage of indirect identifiers is higher than that of direct identifiers.}
\label{fig:mlmdirect}

\end{figure}

A model can leak information by predicting or regurgitating direct or indirect identifiers.
Let us analyze the behavior of \ppmlm for the RoBERTa Large model if we consider only the protection of direct identifiers or indirect identifiers (Figure~\ref{fig:mlmdirect}). 
Results show that only protecting direct
identifiers (i.e., \ppmlmd) provides about the same level of utility while only increasing slightly the privacy compared to \mlm. This small increase in privacy is
due to the small number of distinct direct identifiers. Protecting only the indirect identifiers (i.e., \ppmlmi), in turn, significantly improves privacy compared to \mlm. Indeed, since the number of distinct indirect identifiers (i.e., a total of 15,263) is significantly greater than the number of direct identifiers (i.e., a total of 8,832), the gain in terms of privacy is much more pronounced.
The remaining difference in terms of privacy compared to \ppmlm corresponds to the lack of protection of direct identifiers.
The utility, on the other
hand, is similar.

\section{Impact of parameter k}
\label{sec:k}

\begin{figure}[t]
    \centering
    \includegraphics[width=0.33\textwidth]{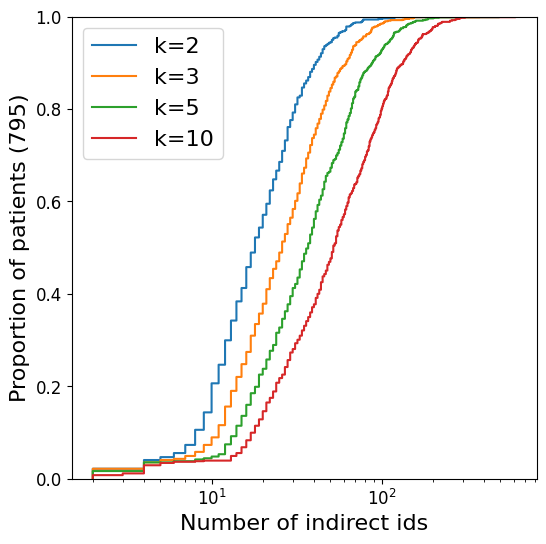}
 
    \caption{Impact of parameter $k$: as $k$ increases, the number of indirect identifiers increases.}
    \label{fig:impactk}
\end{figure}

\begin{figure}[t]
    \centering
    \includegraphics[width=0.48\textwidth]{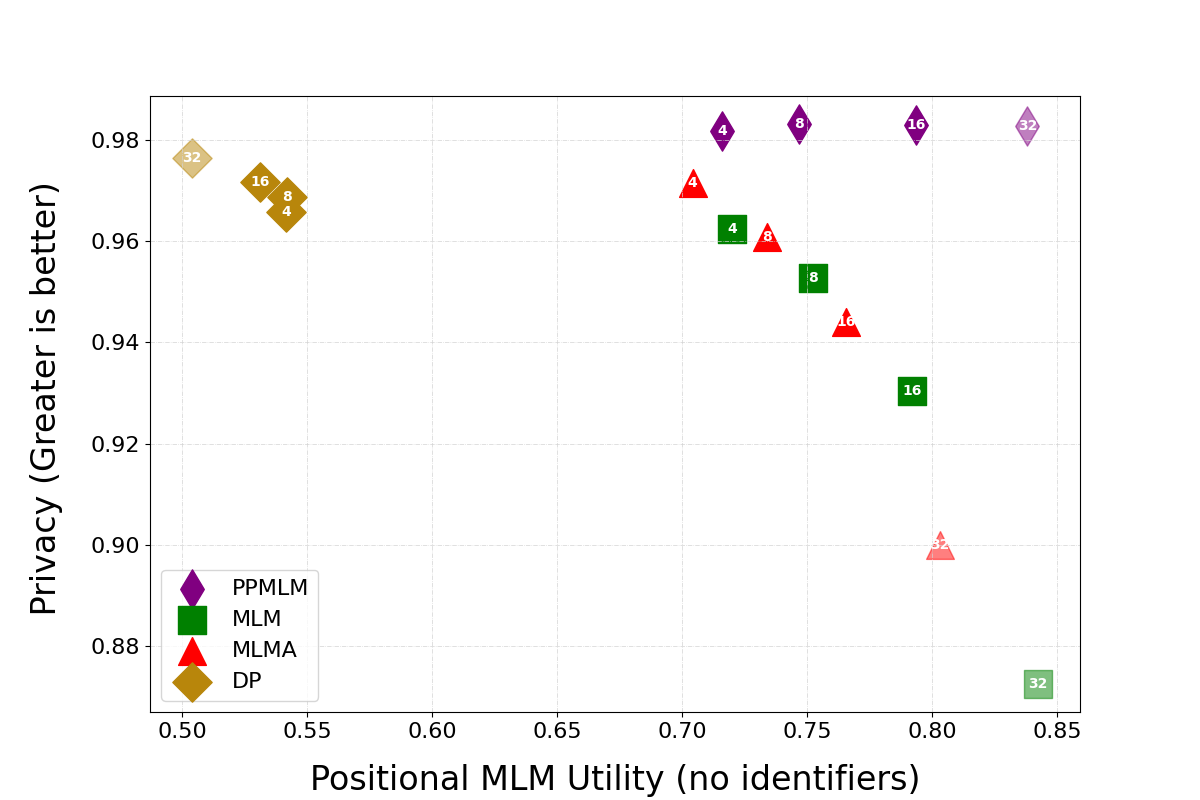}
      \caption{\ppmlm is not impacted by the value of k (RoBERTa LARGE, k=5).}
    \label{fig:mlm-impact-k}
\end{figure}

The preprocessing step of our solution to identify indirectly identifying words is somewhat like implementing k-anonymity with k=2. 
More precisely, this means that a term used by only one individual will be protected (i.e., not memorized by the model). In other words, unprotected terms are, at worst, used in the reports of two patients, and it is difficult for an attacker to easily identify the patient who leaked the information. 
We can easily change this value to a larger value. %
For example, by taking k=3, we ensure that the words that will be predicted by the model are at worst used in the reports of three patients, making the attacker's task even more difficult.
This configurable protection (i.e., the value of k) does not include any additional cost and only consists of navigating the bipartite graph built in the preprocessing step of our solution.
Figure~\ref{fig:impactk} shows the distribution of the number of indirect identifiers for different values of $k$.
The results show that as $k$ increases, the number of indirect identifiers increases.
This trend mechanically increases the number of sensitive words that will not be used as masks to avoid being memorized. In this case, the level of privacy remains unchanged because the additional sensitive words are protected by \ppmlm and will not be predicted by the model. However, this number of words excluded from masks can slightly reduce the model's utility. 
Figure~\ref{fig:mlm-impact-k} depicts the utility and privacy tradeoff for RoBERTa Large model when k=5.
The results show that the trend is similar to that observed with k=1 (Figure~\ref{fig:mlmprivacy}) and that \ppmlm is not impacted by the value of k. We show that the privacy gap is slightly more pronounced for \mlm and \mlma due to a larger number of words identified as sensitive (i.e., k=5 increases the number of indirect identifiers compared to k=1).


\section{Considering n-grams}
\label{sec:impactngrams}

\begin{figure}[t]
    \centering
    \includegraphics[width=0.33\textwidth]{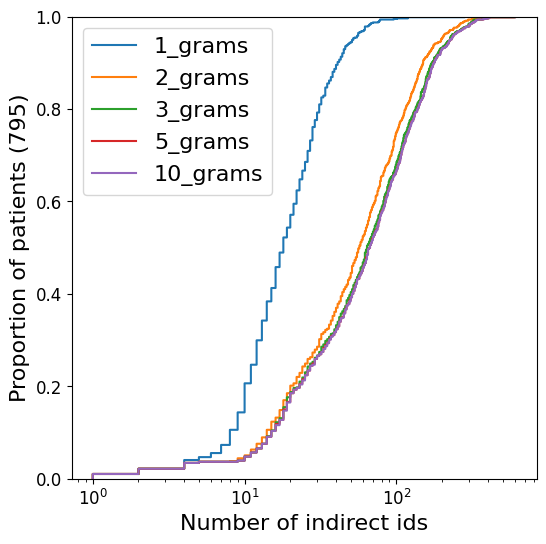}
 
    \caption{Distribution of the number of indirect identifiers by considering n-grams as potential identifiers: the number of indirect identifiers does not increase  after n=2.}
    \label{fig:ngrams}
\end{figure}

The proposed solution only considers isolated words as indirect identifiers. However, these indirect identifiers can be composed of multiple terms (i.e., n-grams of variable size n). In other words, the value for n only applies to indirect identifiers. For direct identifiers, the number of words depends on the nature of the identifier returned by the NER; for example, a date or an address will consist of several contiguous words, all of which are protected.

\begin{figure}[t]
    \centering
    \includegraphics[width=0.48\textwidth]{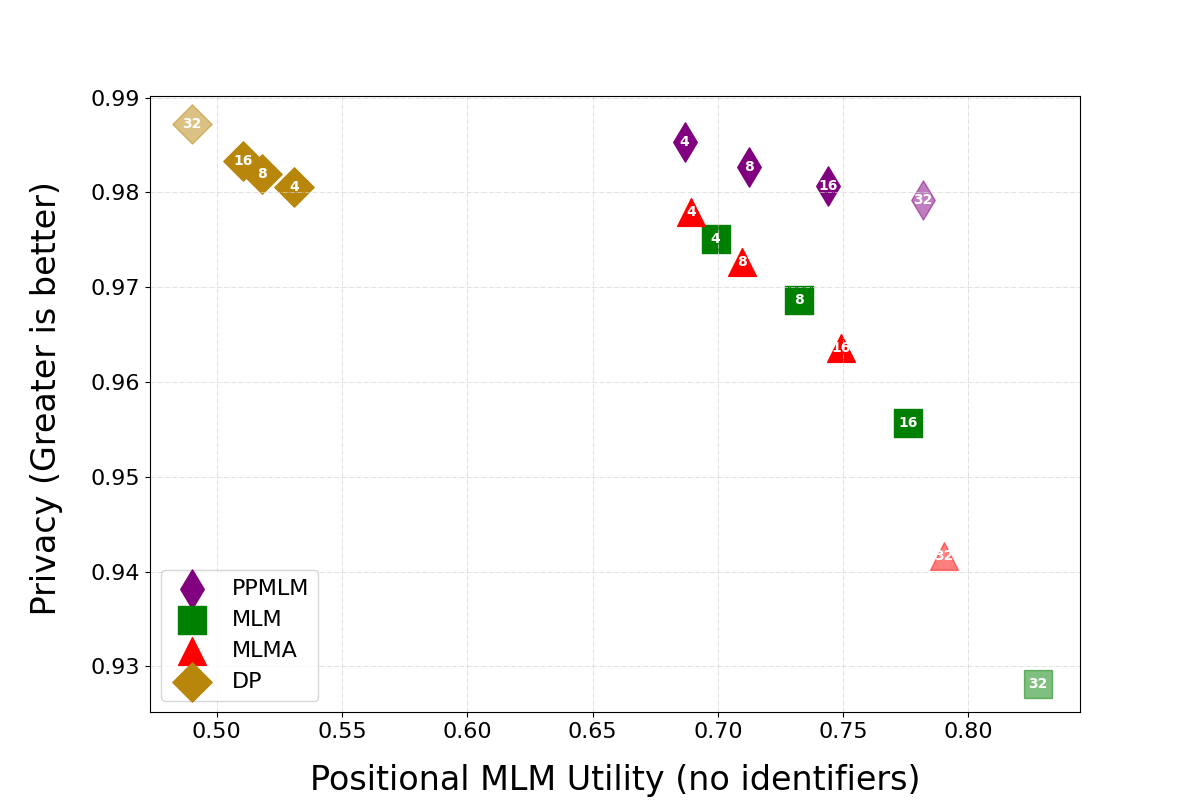}
 \label{fig:mlmprivacyN2C2}
 
     \caption{Impact of n on the privacy and utility tradeoff (RoBERTa
LARGE, n=3): }
    \label{fig:mlm-impact-n}
\end{figure}

Extending our solution to consider these n-grams as identifiers involves generating a bipartite graph accounting for them.
We first analyze the dataset to quantify the number of n-grams detected as indirect identifiers (i.e., unique to a patient).
Figure~\ref{fig:ngrams} illustrates the distribution of the number of n-grams as indirect identifiers per patient for different values of n.
Results show that, on average, the number of indirect identifiers doubles if we consider n from 1 (i.e., a word unique to a patient) to 2 (i.e., a sequence of two words unique to a patient). Considering n greater than 2 does not significantly increase the number of indirect identifiers.
Figure~\ref{fig:mlm-impact-n} depicts the utility and privacy tradeoff for
RoBERTa Large model when n=3. Results show that the trend is similar to that observed with n=1 (Figure~\ref{fig:mlmprivacy}).

\begin{figure}[t]
    \centering
    \includegraphics[width=0.35\textwidth]{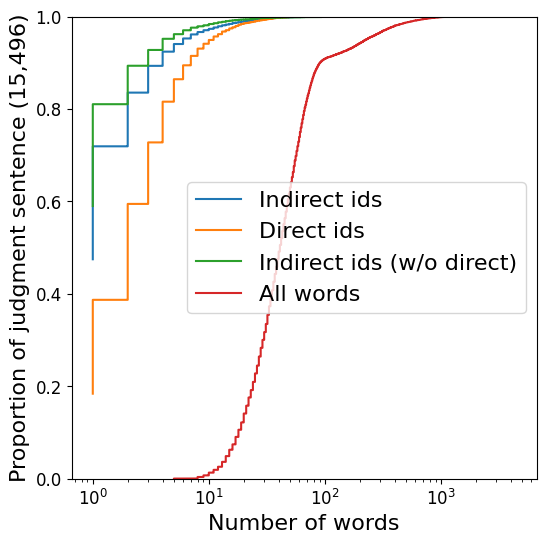} 
     \caption{Cumulative distribution of the number of identifiers and words per judgment sentence: fewer identifiers (direct and indirect) than in \nc, but the same trend.}
    \label{fig:cdf_ind_ids_legal}
\end{figure}

\begin{figure}[t]
    \centering
    \includegraphics[width=0.35\textwidth]{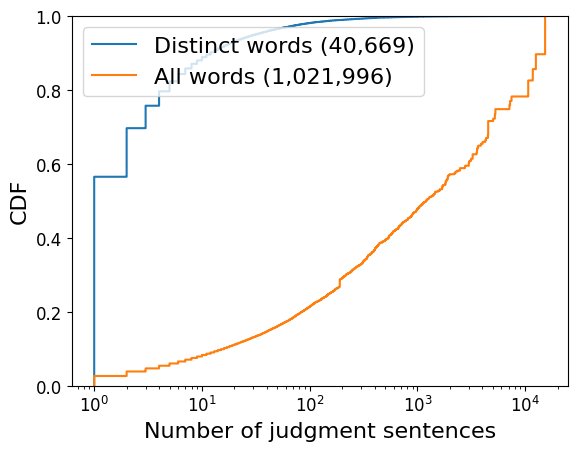} 
     \caption{Cumulative distribution of the number of judgment sentences who share words: 60\% of distinct words are used by only one judgment sentence and represent indirect identifiers, however these words represent only 3\% of the total number of word occurrences in all sentences.}
    \label{fig:prop_ids_legal}
\end{figure}

\section{Assessing the generality}
\label{appendix:legal}

To assess the generality of our solution, we evaluated it with a corpus of legal texts from judgments~\cite{kalamkar2022namedentityrecognitionindian}.  
This corpus gathers 15,496 judgment sentences. 
The NER for this corpus targets 14 legal entities (e.g., Court, Petitioner, Respondent, Judge, Lawyer, Date, ...).
Although the methodology remains the same, the nature of the directly identifying entities varies.
First, we analyze the distribution of direct and indirect identifiers for each judgment in the dataset (Figure~\ref{fig:cdf_ind_ids_legal}).
We observe that the number of identifiers is mostly less than that of the \nc dataset. This is due to a judgment sentence length smaller than the size of the medial reports. However, the number of judgment sentences is much greater than the number of patients (15,496 judgment sentences versus 795 patients).
The trends remain similar: the majority of judgment sentences include identifiers, direct identifiers are approximately twice as numerous as indirect identifiers, and some indirect identifiers are also direct identifiers.
Then, we analyze the cumulative distribution of the number of judgment sentences which share words (distinct words and the total number of words by including repetitions) in their text (Figure~\ref{fig:prop_ids_legal}).
The distribution remains similar to that of \nc and shows that 60\% of distinct words in the corpus are used by only one judgment sentence and therefore represent indirect identifiers.
However, similar to the other dataset, the use of these words represents only a small percentage (around 3\%) of the total number of word occurrences in all judgment sentences. In other words, the 40\% of words used in multiple sentences account for more than 97\% of the total number of words used.

 

\begin{figure}[t]
    \centering
    \includegraphics[width=0.48\textwidth]{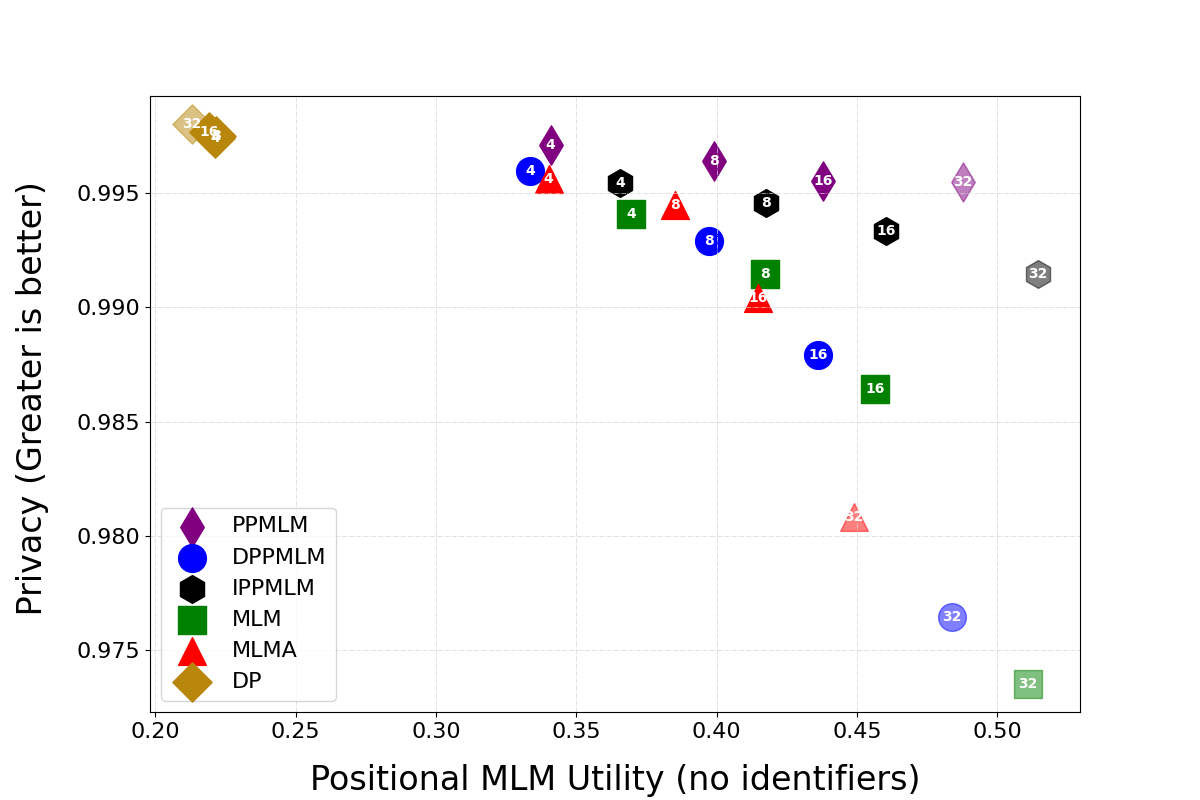}
 
     \caption{Utility and privacy evaluation with the corpus of legal texts (BERT Base-Cased): similar to \nc, by avoiding memorizing both direct and indirect identifiers during model specialization, our \ppmlm solution offers the best utility and privacy tradeoff.}
    \label{fig:mlmprivacy_legal}
\end{figure}

We then evaluate the utility and privacy tradeoff for our masked 
language modeling scheme using this corpus of legal texts.
Figure~\ref{fig:mlmprivacy_legal} 
depicts the results for \ppmlm on the BERT model. 
The results show the same trend as observed with the \nc dataset in both cases: an important decrease in privacy with a solution without protection, an important decrease in utility for the solution based on DP, and a better compromise for our solution.

\end{document}